\def\year{2019}\relax
\documentclass[letterpaper]{article} 
\usepackage{aaai19}  
\usepackage{times}  
\usepackage{helvet}  
\usepackage{courier}  
\usepackage{url}  
\usepackage{graphicx}  
\usepackage{amsmath,amssymb}
\usepackage{multirow}
\usepackage{subfigure}
\usepackage{color}

\newcommand{\etal}{\emph{et al.}\ }
\newcommand{\citep}{\cite}
\newcommand{\ssc}{\sigma} 
\newcommand{\eg}{\textit{e.g.}\ } 
\newcommand{\ie}{\textit{i.e.}\ }

\newcommand{\Earlyfusion}{{LSTM}}
\newcommand{\Latefusion}{{VE}} 
\newcommand{\queryproposal}[1]{{#1+QSPN}}
\newcommand{\capt}[1]{{#1+Cap}}

\begin{document}
%
\title{Multilevel Language and Vision Integration for Text-to-Clip Retrieval}
\author{Huijuan Xu\footnotemark[2] \hspace{4mm}  Kun He\footnotemark[2] \hspace{4mm}  Bryan A. Plummer\footnotemark[2] \hspace{4mm}  Leonid Sigal\footnotemark[3] \hspace{4mm}  Stan Sclaroff\footnotemark[2] \hspace{4mm}  Kate Saenko\footnotemark[2]\\
\footnotemark[2]Boston University, \footnotemark[3]University of British Columbia\\
\footnotesize\footnotemark[2]\texttt{\{hxu, hekun, bplum, sclaroff, saenko\}@bu.edu}, \footnotemark[3]\texttt{lsigal@cs.ubc.ca}\\
}
\maketitle
\begin{abstract}

We address the problem of text-based activity retrieval in video. Given a sentence describing an activity, our task is to retrieve matching clips from an untrimmed video. 
To capture the inherent structures present in both text and video, we introduce a multilevel  model that integrates vision and language features earlier and more tightly than prior work.
First, we inject text features early on when generating clip proposals, to help eliminate unlikely clips and thus speed up processing and boost performance. Second, to learn a fine-grained similarity metric for retrieval, we use visual features to modulate the processing of query sentences at the word level in a recurrent neural network. 
A multi-task loss is also employed by adding query re-generation as an auxiliary task.
Our approach significantly outperforms prior work on two challenging benchmarks: Charades-STA and ActivityNet Captions.

\end{abstract}

\section{Introduction}
\label{sec:intro}

\begin{figure}[t]
\includegraphics[width=.88\linewidth]{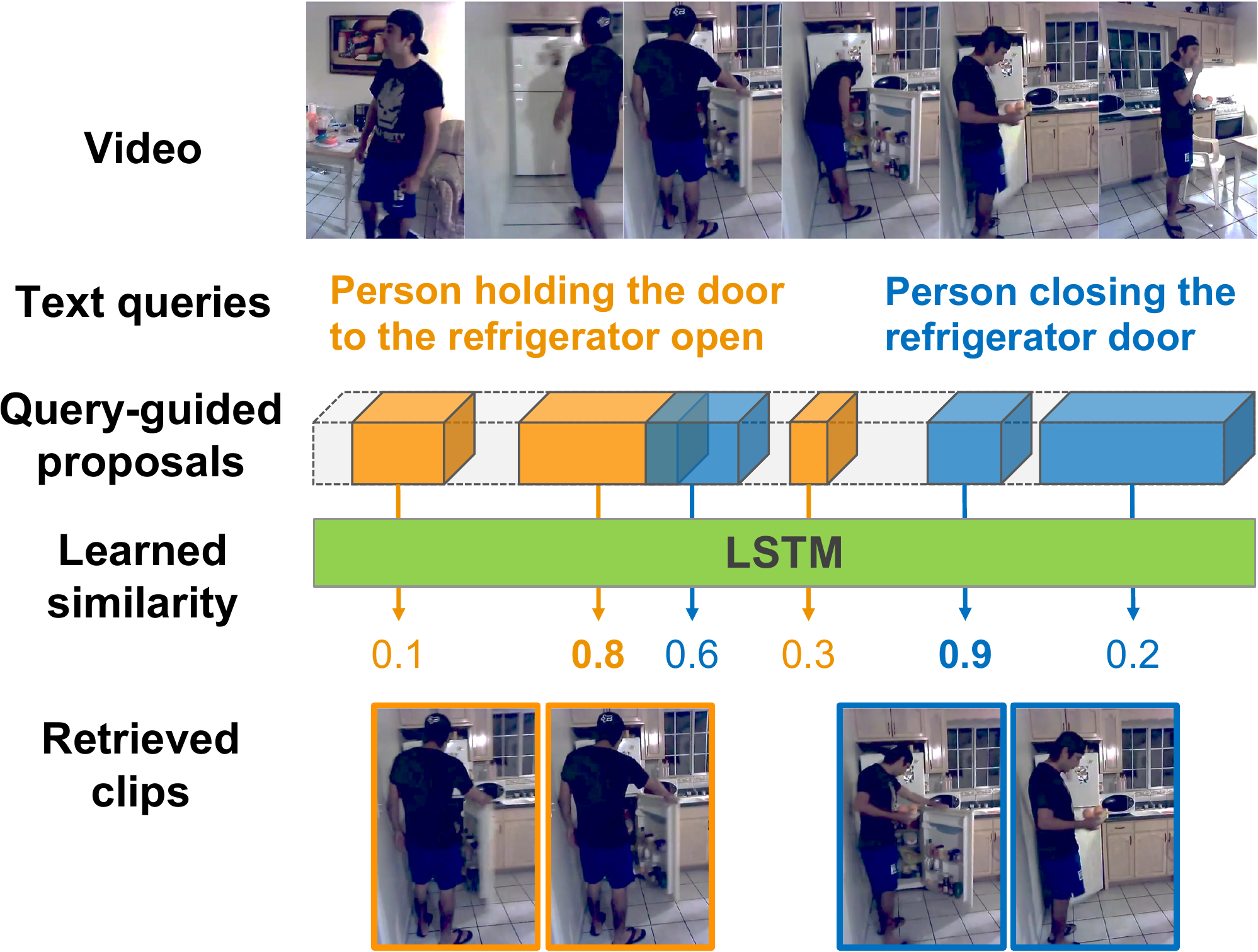}
\caption{
\emph{Text-to-clip}
is a complicated task that can involve objects, their interactions, and activities.
In this example, the model needs to understand the objects being referred to (person and refrigerator), disambiguate between the verbs (holding open \emph{vs.} closing), and perform temporal localization based on the movement of the refrigerator door.
For this task, we propose a \emph{multilevel} model to tightly integrate language and vision features in its retrieval pipeline: 1) query-guided video clip proposals, 2)  learned similarity measure with a recurrent LSTM.
}
\label{fig:overview}
\end{figure}

Temporal localization of events or activities of interest is a key problem in computer vision. Recently there has been increased interest in specifying the queries using natural language rather than only supporting a predefined set of actions or events \cite{TALL,didemo}.
In this paper, we focus on the task of retrieving temporal segments in untrimmed video through natural language queries, or simply, \emph{text-to-clip}.
Solving this task requires understanding the nuances of natural language and video contents. 
For example, in Figure~\ref{fig:overview}, although the two queries talk about the same objects,  their verbs make the key difference,  and the temporal ordering of video frames can give an important cue.

Existing methods for solving cross-modal retrieval tasks 
typically learn embedding functions to project data from different modalities into a common vector embedding space. In this common space retrieval is performed using standard similarity metrics, such as Euclidean distance.
However, for \emph{text-to-clip}, such wholistic representations of sentences and video clips make it difficult to leverage fine-grained structures, such as the ordering of words and frames.
Also, the embeddings are usually independent of each other,  
making it impossible to use input from one modality to modulate the processing of another.

To better model the structured information present in 
both modalities,
we propose a novel \emph{multilevel} approach for integrating vision and language features for the {text-to-clip} task.  In the first level, we inject language features at the temporal segment proposal stage. Inspired by attention mechanisms in Visual Question Answering (VQA)~\cite{Anderson_2018_CVPR,lu2016hierarchical,shih2016wtl,xu2016ask}, we re-weight the video features used to propose temporal segments by their similarity to the encoding of the input query.  
This enables our model to select clips that are more relevant to  queries as candidates for further processing.
The video feature weights can be pooled over a set of queries which helps to reduce computational costs when performing multiple queries on a single video.

The second level of integration happens when computing the similarity between the text query and video clips. 
We learn a Long Short-Term Memory (LSTM) model that processes the query sentence word-by-word, conditioned on the visual feature embedding of a candidate clip in each step, and produces a nonlinear similarity score.
This can be seen as an \emph{early fusion} of vision and language, as opposed to the \emph{late fusion} in vector embedding   approaches.
Also, our model has the freedom to associate each word in the query sentence with a potentially different part of the visual feature when computing the similarity, which is not possible in independent vector embeddings.
Additionally, we train our model with a multi-task loss, adding \emph{clip-to-text} as an auxiliary task, which is also known as dense video captioning \cite{DenseCapVid}. We show that learning a shared representation for both tasks improves retrieval performance.
An overview of our approach is provided in Figure~\ref{fig:overview}.

Our approach differs from prior work in two important ways.  First, our model learns to propose temporal segments conditioned on the query, unlike exiting solutions for this task that employ sliding windows or hand-crafted heuristics \cite{didemo,TALL} that do not consider the query at all.  
We generalize the R-C3D model~\cite{R-C3D} to create query-specific proposals.
Second, we improve upon vector embedding approaches 
that pool the hidden states of a recurrent neural network to obtain sentence embeddings \cite{didemo,TALL,aytar2017see}.  
These models would have to anticipate what words are important without having access to the visual data.  
Our approach addresses this drawback by integrating visual features while processing sentence queries  at the word level.

To summarize our contributions, for the text-to-clip retrieval task, in this paper we
\begin{itemize}
\item incorporate query embeddings into the segment proposal network to generate query-guided proposals;
\item take an early fusion approach and learn an LSTM  to model fine-grained similarity between query sentences and video clips;
\item leverage captioning as an auxiliary task to learn better shared feature representations.
\end{itemize} 

We conduct extensive evaluation and ablation studies on  two challenging benchmarks: Charades-STA~\citep{TALL} and ActivityNet Captions~\cite{DenseCapVid}. 
Our full model achieves state-of-the-art  performance. Code is released for public use\footnote{\tiny \url{https://github.com/VisionLearningGroup/Text-to-Clip_Retrieval}}.

\section{Related Work}
\label{sec:related}

\noindent\textbf{Temporal Activity Detection:}
Temporal activity detection aims at predicting the start and end times of the activities within untrimmed videos. 
Early approaches~\cite{shou2016temporal} use sliding windows to generate segments and subsequently classify them,  which is computationally inefficient and constrains the granularity of detection.
Later in \cite{Singh2016cvpr,yeung2016end} temporal localization is obtained by modeling the evolution of activities using Recurrent Neural Networks (RNNs) and predicting activity labels or activity segments at each time step.
R-C3D~\cite{R-C3D} adapts the proposal and classification pipeline from object detection~\cite{faster-rcnn} to perform activity detection using 3D convolutions~\cite{C3D} and 3D Region of Interest pooling. 
SSAD~\cite{lin2017single} performs single-shot temporal activity detection following the one-stage object detection method SSD~\cite{liu2016ssd}. 

Instead of treating activities as distinct classes and using a discrete and fixed vocabulary of class labels, language queries in video language localization task can express more semantic meaning. 
In this paper, we use a proposal-based pipeline to solve the video language localization task, and adopt the proposal generation technique of R-C3D.

\smallskip
\noindent\textbf{Vision and Language:}
In this paper our task is to locate the visual events that match a query sentence in a video. 
There are two main types of approaches to solve such cross-modal retrieval tasks: early fusion and late fusion.
The late fusion approach embeds different modalities into a common embedding space, and then measure their similarity.
These approaches are not restricted to vision and language, but can be applied across modalities such as image, video, text, and sound 
\cite{arandjelovic2017objects,aytar2017see,vendrov2015order}.  Early fusion approaches combine the features from each modality at an earlier stage~\cite{ma2015multimodal,wang2018learning,yu2017end} and  predicts similarity scores directly based on the fused feature representation. 
\cite{NIPS2017_vries} argues against the dominant late-fusion pipeline, where linguistic inputs are mostly processed independently, and shows that modulating visual representations with language at earlier levels improves visual question answering.

For the text-to-clip task considered in this paper, existing models~\cite{TALL,didemo} perform late fusion and embed entire query sentences to vectors.
However, as we argued in the introduction, this tends to lose important information about fine-grained structures.
We propose a novel model that performs early fusion of the video and query features, combining them at the word level, and we compare it with the sentence-level fusion approach.

Other typical vision-language tasks include image/video captioning~\cite{donahue2015long,venugopalan2015translating,venugopalan15iccv,vinyals2015show,yao2015describing,xu2015multi} and visual question answering (VQA)~\cite{xiong2016dynamic,yang2016stacked}.
We note that these tasks are rarely isolated and often influence each other.
For example, image captioning can be solved as a retrieval task~\cite{fang2015captions}.
Also, there is recent research that suggests that VQA can be leveraged to benefit the image-caption retrieval task~\cite{lin2016leveraging}.
Our proposed multi-task formulation, which uses captioning as an auxiliary task, is partly motivated by these observations.

\smallskip
\noindent\textbf{Localization-based Cross-modal Tasks:}
Several vision-language tasks also share the need for a localization component. 
In the dense captioning task, models need to localize interesting events in images~\cite{densecap} or videos~\cite{DenseCapVid,xu2018joint} and provide textual descriptions.
Recently, the task of grounding text in images~\cite{hu2016natural,rohrbach2016grounding} has been extended into videos, which introduces the task of retrieving video segments using language queries \cite{didemo,TALL}.  These visual grounding approaches have included models which reconstruct the text query (\eg\cite{Javed-arxiv-2018,rohrbach2016grounding}), which we also take advantage of.
We note that the localization mechanisms in \cite{didemo,TALL} are either inefficient (sliding-window based) or inflexible (hard-coded). 
In contrast with these approaches, we adopt learned segment proposals in our pipeline.

\section{Approach}
\label{sec:method}

We propose a novel approach for temporal activity localization and retrieval based on input language queries, or the text-to-clip task. 
Our key idea is to integrate language and vision more closely before computing a match, using an early fusion scheme, query-specific proposals, and a multi-task formulation that re-generates the caption.

We first define the cross-modal retrieval problem we are solving.
Given an untrimmed video $V$ and a sentence query $S$, the goal is to retrieve a temporal segment (clip) $R$ in $V$ that best corresponds to $S$.
In other words, we learn a mapping $\mathcal{F}_{\text{RET}}: (V, S)\mapsto R$.
At training time, we are given a set of annotated videos $\{V_1,V_2,\ldots,V_N\}$. 
For each video $V_i$, its annotation is a set of matching sentence-clip pairs $A_i=\{(S_{ij},R_{ij})\}_{j=1}^{n_i}$, where  $S_{ij}$ is a sentence, and
clip $R_{ij}=(t^0_{ij},t^1_{ij})$ is represented as a pair of timestamps that define its start and end.
We tackle the retrieval problem through learning a similarity score $\ssc(S,R)$ that measures how well $S$ and $R$ match each other.
At test time, given $V$ and $S$, the retrieval problem is formulated as
\begin{align}
R^*=\arg\max_{R\in V}~\ssc(R,S).
\end{align}

The remainder of this section is organized as follows. First, we introduce our query-guided temporal segment proposal network.  
Then, we detail how we learn a fine-grained similarity model for retrieval.
Finally, we describe the multi-task loss function which combines the retrieval loss with an auxiliary captioning loss.

\subsection{Query-Guided Segment Proposal Network}
\label{sec:SPN}

For unconstrained localization in videos, it is important to generate variable-length candidate temporal segments for further processing. 
Instead of using handcrafted heuristics or computationally expensive multiscale sliding windows, 
we employ a learned segment proposal network (SPN), similar to the one used in R-C3D \cite{R-C3D} for action localization.
The SPN first encodes all frames in an input video using a 3D convolutional network (C3D) \cite{C3D}.
Then, variable-length segment proposals are obtained by predicting  relative offsets to a set of predefined  anchor segments.
The proposal features are generated by 3D Region of Interest Pooling.  

\begin{figure}[t]
\centering
\includegraphics[width=0.8\linewidth,trim=1cm 5cm 3.5cm 3.5cm,clip]{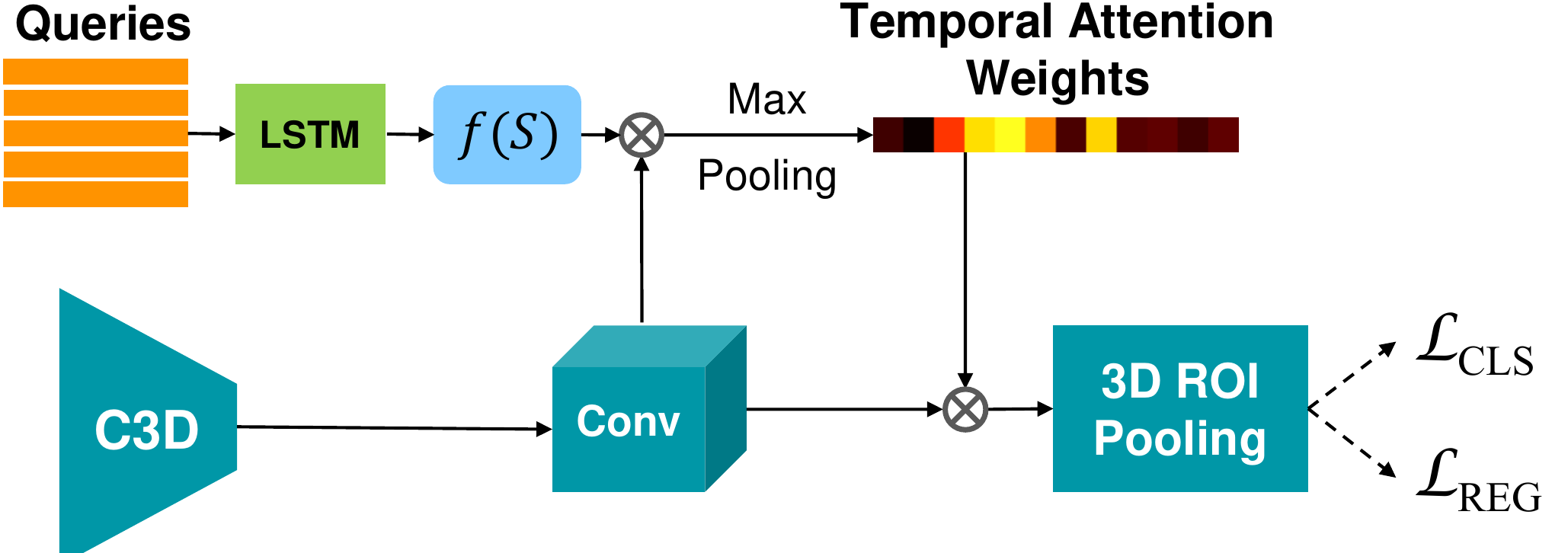}
\caption{
The structure of the query-guided Segment Proposal Network.
Query embeddings are used to derive attention weights and re-weight the video convolutional features before generating proposals.
The SPN is trained with a combination of classification loss $\mathcal{L}_\text{CLS}$ and regression loss $\mathcal{L}_\text{REG}$.
}
\label{fig:videoRetrieval_query_spn}
\end{figure}

We note that the original SPN from R-C3D aims at finding ``anything interesting'' for unconstrained action localization. However,  in the text-to-clip task where a query is specified, the search space can be further reduced by only generating proposals that are relevant to the query.
We thus develop a \emph{query-guided} SPN.
The basic idea is to use the feature representation of the query to modulate the SPN and attend to the relevant temporal regions, in a way that is similar to attention mechanisms in Visual Question Answering (VQA) \cite{lu2016hierarchical,xu2016ask,shih2016wtl}.

Figure~\ref{fig:videoRetrieval_query_spn} shows the structure of the query-guided SPN.
The original SPN is shown at the bottom.
Each query sentence $S$ is embedded into a feature vector $f(S)$ by pooling the hidden states of a sentence embedding LSTM (described in the next subsection).
Then, for each temporal location, an attention weight is computed by taking the inner product of the video features and $f(S)$, and passing through the $\tanh$ activation.
The attention weight at each temporal location is multiplied with the corresponding video features across all channels. When there are multiple queries, we max-pool the weights over the query dimension.

\subsection{Early Fusion Retrieval Model}
\label{sec:LRM}

The output from the segment proposal network is a set of temporal segments likely to contain the relevant activity, along with their pooled C3D features.
We next need a retrieval model to find the segment that best matches a query.

As shown in Fig.~\ref{fig:LSTM}, the pooled C3D features of a clip, along with the query sentence, are fed as input to a two-layer LSTM. 
The first layer of the LSTM processes the words in the sentence. 
In the second layer, the visual feature embedding is used as input at each step, along with hidden states from the sentence embedding LSTM.
The final hidden state is passed through additional layers to predict a scalar similarity value.
We note that while our approach does increase the number of learnable parameters in the model, it also brings additional structure into the similarity metric. This comes from each word in the sentence now being able to interact with the visual features, enabling the model to learn a potentially different way to associate each word and the visual features. 
We do not explicitly use attention mechanisms to enforce such behavior, but instead let the LSTM learn in a data-driven manner.
Our retrieval model is an \emph{early fusion} model, where the processing of visual features and language features are intertwined rather than isolated.

To train the retrieval model, we use a triplet-based retrieval loss, also called pairwise ranking loss~\cite{hullermeier2008label}, which has shown good performance in metric learning tasks~\cite{hoffer2015deep,facenet}. 
Specifically, we take triplets of the form $(S,R, R')$ where $(S,R)$ is a matching sentence-clip pair, and $R'$ is a negative clip 
that does not match $S$.
Note that $R'$ can either come from the same video as $R$ with a sufficiently low overlap, or from a different video.
The loss encourages the similarity score between the matching pair, $\ssc(S,R)$, to be greater than $\ssc(S,R')$ by some margin $\eta >0$:
\begin{align}
\mathcal{L}_{\text{RET}} = \sum_{(S,R,R')}
\max\{0, \eta+\ssc(S,R')-\ssc(S,R)\}.
\end{align}
In our model, $\ssc(S,R)$ is directly predicted by the LSTM rather than using some generic measure, like cosine similarity, as done in prior work (\eg~\cite{didemo}).

\subsection{Multi-Task Loss}
\label{sec:MT}

\begin{figure}[t]
\centering
\includegraphics[width=.85\linewidth,trim=2.cm 3.2cm 4cm 3.8cm,clip]{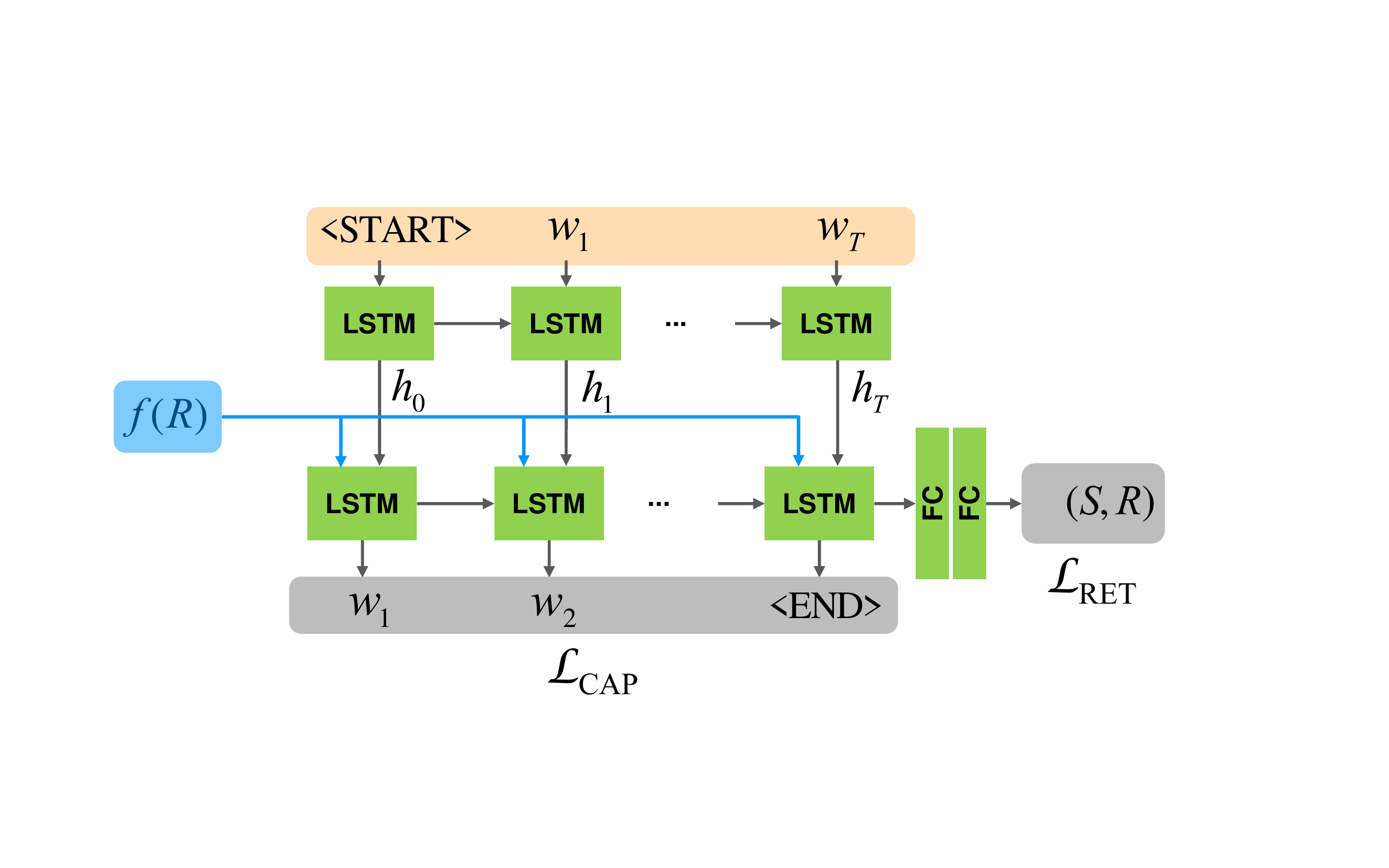}
\caption{Our early fusion retrieval model with multi-task loss, instantiated as a two-layer LSTM.
The first layer embeds the sentence query $S$, while the second layer takes both $h_0$ and the visual feature of a clip $f(R)$ as (separate) inputs, and predicts a nonlinear similarity score $\sigma(S,R)$ that is supervised by the retrieval loss $\mathcal{L}_\text{RET}$.
Additionally, we add a captioning loss $\mathcal{L}_\text{CAP}$ to enforce the LSTM to re-generate the query sentence,
resulting in improved retrieval performance.
}
\label{fig:LSTM}
\end{figure}

After defining the retrieval model, we now seek to gain additional benefit from training on closely related tasks.  Specifically, we add a captioning loss which can act as a verification step for our model, \ie we should be able to re-generate the query sentence from the retrieved video clip. 
Captioning has also proven to improve performance on image-based multimodal retrieval tasks \cite{rohrbach2016grounding}.  
Moreover, it is observed \citep{Ramanishka2017cvpr} that captioning models can implicitly learn features and attention mechanisms to associate spatiotemporal regions to words in the captions.
As for implementation, the paired sentence-clip annotation format in the text-to-clip task allows us to 
easily add captioning capabilities to our LSTM model.

As shown in Figure~\ref{fig:LSTM}, we require the second layer of our LSTM to re-generate the input query sentence, 
conditioned on the proposal's visual features $f(R)$ at each step.
When generating word $w_t$ at step $t$, the hidden state from the previous step in the sentence embedding LSTM, $h_{t-1}^{(1)}$, is used as input. 
We use a standard captioning loss that maximizes the normalized log likelihood of the words generated at all $T$ unrolled time steps,
over all $K$ ground truth matching sentence-clip pairs: 
\begin{align}
\mathcal{L}_{\text{CAP}} =  -\frac{1}{KT}\sum_{k=1}^K \sum_{t=1}^{T_k} \log P(w^{k}_{t}| f(R), h_{t-1}^{(2)} ,w^{k}_{1},...,w^{k}_{t-1}).
\label{eq:loss}
\end{align}

\subsection{Implementation Details}
\label{sec:implementation}

Our multi-task model uses weighting parameter $\lambda$ to optimize a combination of retrieval loss and captioning loss:
\begin{align}
\mathcal{L}=\mathcal{L}_{\text{RET}}+\lambda \mathcal{L}_{\text{CAP}}.
\label{eq:loss_weight}
\end{align}
We choose $\lambda=0.5$ through cross-validation. 
The margin parameter $\eta$  is set to 0.2 in the retrieval loss $\mathcal{L}_{\text{RET}}$. 
During training, each minibatch contains 32 matching sentence-clip pairs sampled from the training set, which are then used to construct triplets.
We use the Adam optimizer \cite{adam} with learning rate 0.001 and early stopping on the validation set, for  30 epochs in total.

For the sentence embedding LSTM (first layer), we use \texttt{word2vec} \cite{word2vec} as the input word representation.
The word embeddings are 300-dimensional, and trained from scratch on each dataset. 
The hidden state size of the LSTM is set to 512.
The size of common embedding space in the late fusion retrieval model is 1024.

The similarity score for a sentence-clip pair is produced by taking the hidden state corresponding to the last word in the second layer of the LSTM,  and passing it through two fully-connected (FC) layers followed by a sigmoid activation to produce a scalar value $\ssc$, as shown in Fig.~\ref{fig:LSTM}. The two FC layers reduce the dimensionality from 512 to 64 to 1.

At test time, retrieving clips in untrimmed videos involves searching over all possible segments.
Candidate proposal segments generated from the proposal network are filtered by non-maximum suppression with threshold 0.7, and the top 100 proposals in each video are used for retrieval.

\section{Experiments}
\label{sec:experiment}

\begin{table*}[t]
\renewcommand{\arraystretch}{1.2}
\centering

\begin{tabular}{l||ccc|ccc|ccc}
\hline
\multirow{2}{*}{Methods} 
& \multicolumn{3}{|c}{tIoU=0.3} 
& \multicolumn{3}{|c}{tIoU=0.5}  
& \multicolumn{3}{|c}{tIoU=0.7}  \\
& \multirow{1}{2.4em}{\,R@1} & \multirow{1}{*}{R@5} & \multirow{1}{2.5em}{R@10} 
& \multirow{1}{2.4em}{\,R@1} & \multirow{1}{*}{R@5} & \multirow{1}{2.5em}{R@10} 
& \multirow{1}{2.4em}{\,R@1} & \multirow{1}{*}{R@5} & \multirow{1}{2.5em}{R@10} \\
\hline
Random \cite{TALL}   
          & -- & -- & --
          & 8.5 & 37.1 & --
          & 3.0 & 14.1 & --\\
          
CTRL(reg-np) \cite{TALL}           
          & -- & -- & --
          & 23.6 & 58.9 & --
          & 8.9 & 29.5 & --\\
\hline
\Latefusion 
          & 43.9 & 83.5 & 89.7 
          & 26.3 & 63.9 & 78.2  
          & 10.9 & 35.6 & 50.5 \\
\Earlyfusion   
          & 51.6 & 95.5 & 99.0
          & 32.8 & 76.3 & 92.5
          & 14.0 & 43.2 & 60.7 \\
\capt{\Earlyfusion} 
          & 52.3 & 95.3 & \textbf{99.2}
          & 34.4 & 77.0 & 92.5  
          & 15.6 & 44.9 & 61.4 \\
\queryproposal{\Earlyfusion}
          & 54.1 & \textbf{95.8} & \textbf{99.2}
          & 35.3 & 77.8 & 93.5
          & 15.2 & 44.6 & 61.9 \\
\capt{\queryproposal{\Earlyfusion}}
          & \textbf{54.7} & 95.6 & \textbf{99.2} 
          & \textbf{35.6} & \textbf{79.4} & \textbf{93.9}  
          & \textbf{15.8} & \textbf{45.4} & \textbf{62.2} \\
\hline
\end{tabular}
\caption[Text-to-clip retrieval results on Charades-STA dataset]
{Results on the Charades-STA dataset \cite{TALL}. R@$K$ stands for Recall@$K$.
Our early fusion retrieval model \Earlyfusion\ significantly outperforms baselines, while the multi-task and query-guided proposals in model \capt{\queryproposal{\Earlyfusion}}  further improve results.
}
\label{table:charades}
\end{table*}

\begin{table*}[t]
\begin{center} { 
\begin{tabular}{l|c}
\hline
\bf queries: [start:end] sentence & ~~~~~ Expected ~~~~~~~~~~~~\capt{\Earlyfusion}~~~~~~~~~~~~~ \Earlyfusion\ ~~~~~~~~~~~~~~~~~~~~\Latefusion~~~~~~~  \\ \hline 

\begin{tabular}[c]{@{}l@{}l@{}}
video ID: EEVD3 \\  
1. [2.0:7.9] a person is holding the door to the \\~~~~refrigerator open. \\
2. [11.4:16.9] person closing the door.\\
\end{tabular}
&
\begin{tabular}{c} \\ 
\;\;\;
\includegraphics[width=0.6in,height=0.6in]{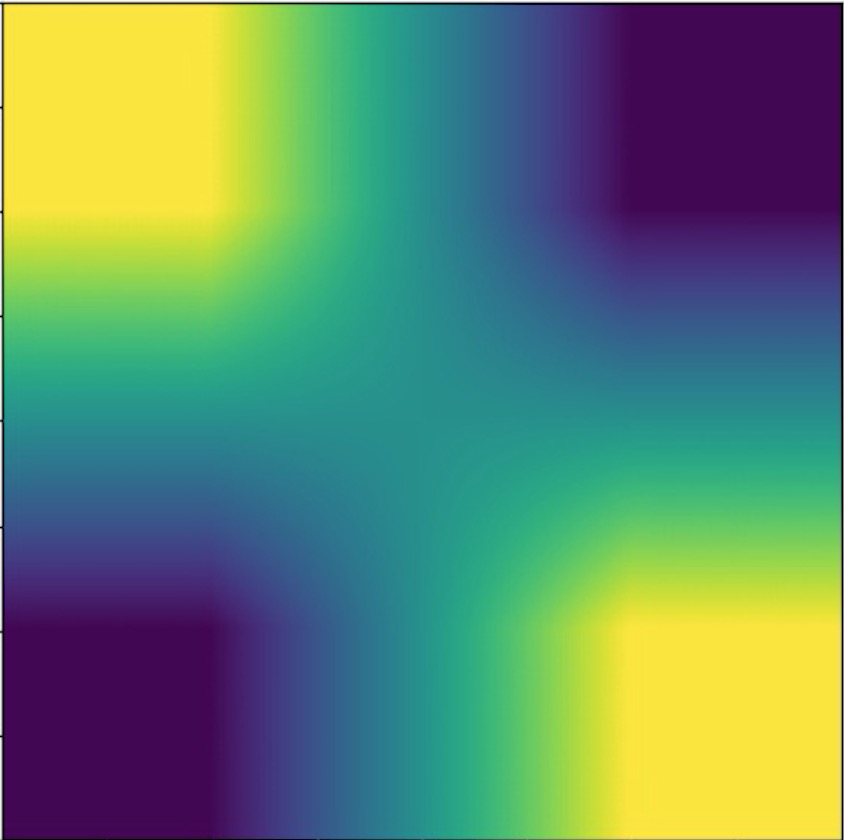} \;\;\;\;\;\;\;\;\;
\includegraphics[width=0.6in,height=0.6in]{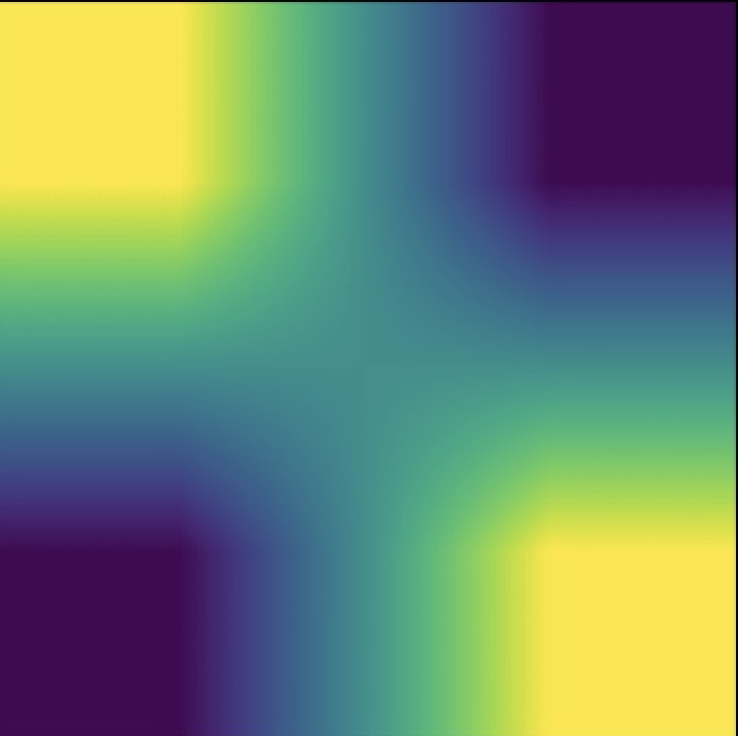}
\;\;\;\;\;\;\;\;\;
\includegraphics[width=0.6in,height=0.6in]{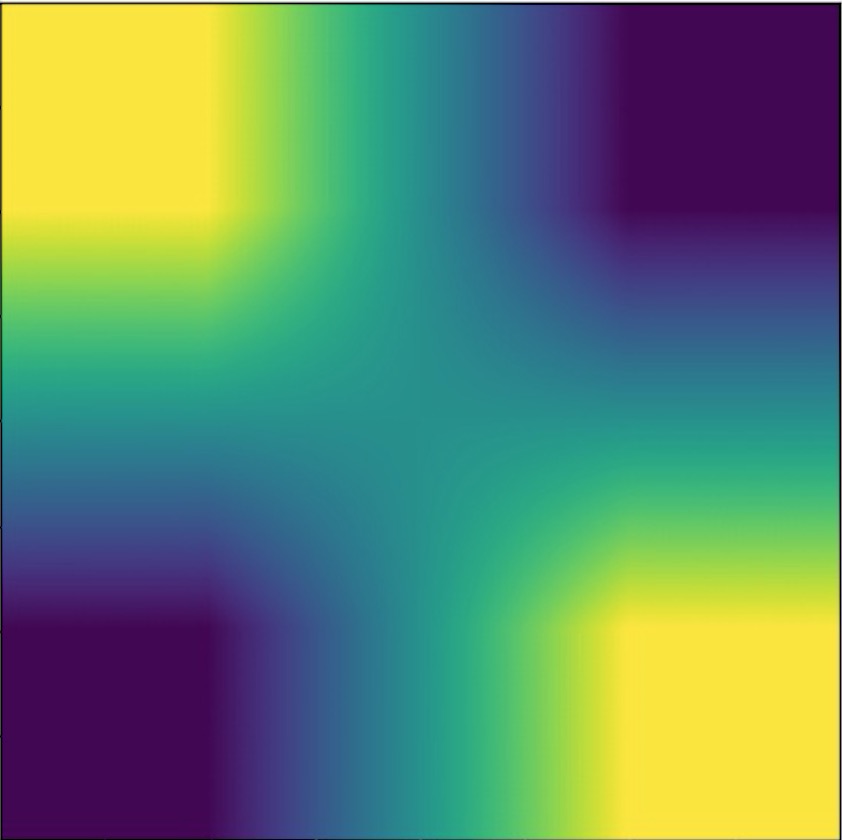}
\;\;\;\;\;\;\;\;\;
\includegraphics[width=0.6in,height=0.6in]{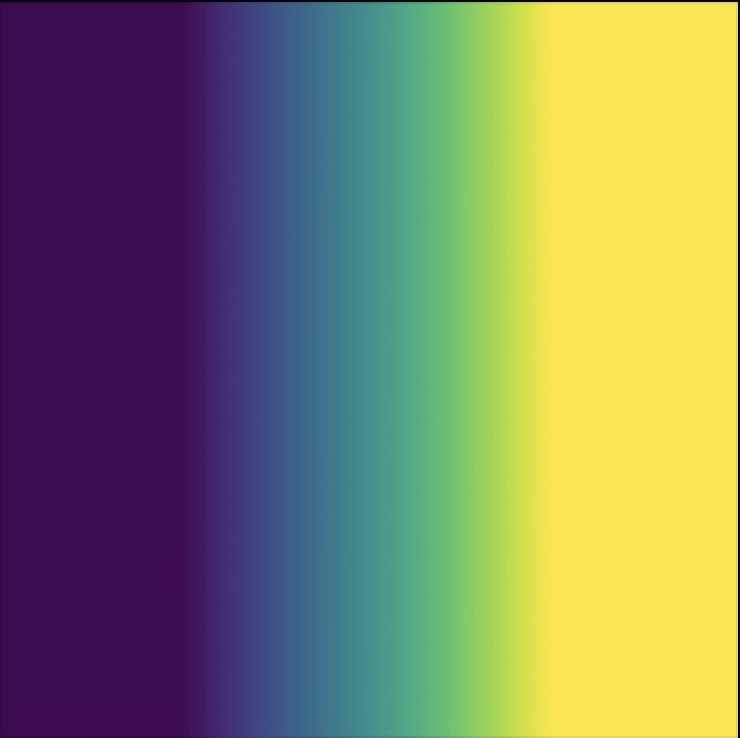}
\end{tabular}
\\  
\hline 

\begin{tabular}[c]{@{}l@{}l@{}}
video ID: 3VT73  \\ 
1. [2.3:11.6]  a person sits down as they read a book.\\
2. [8.4:13.1]  the person throws a book.\\
3. [10.4:15.9] person he takes his cell phone out. \\
\end{tabular}
&
\begin{tabular}{c} \\
\;\;\;
\includegraphics[width=0.6in,height=0.6in]{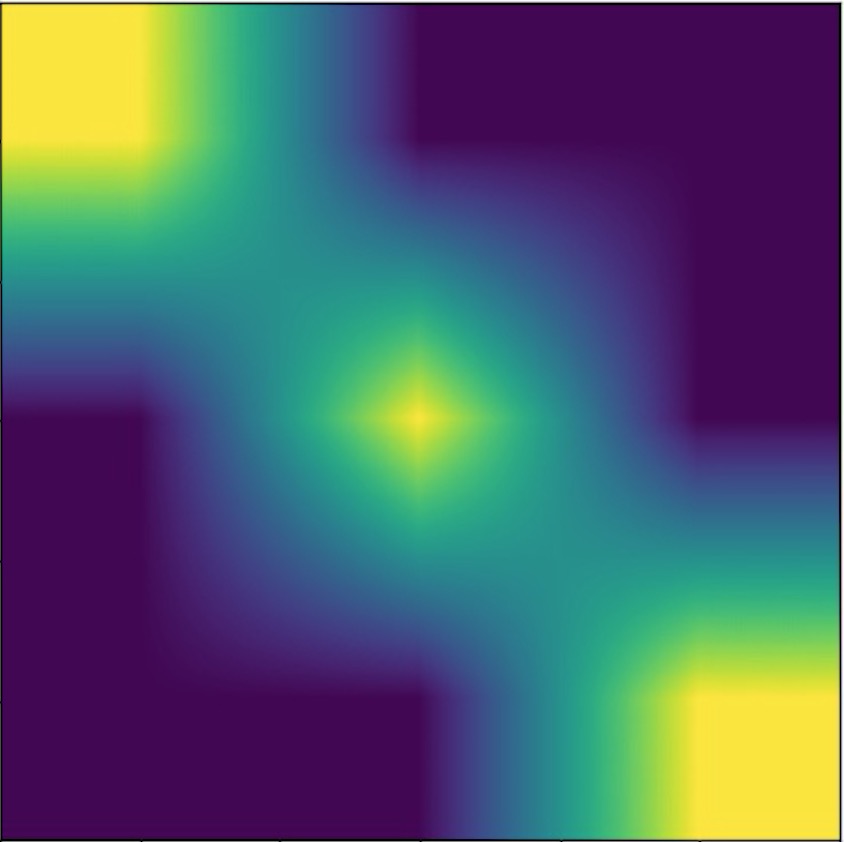} 
\;\;\;\;\;\;\;\;\;
\includegraphics[width=0.6in,height=0.6in]{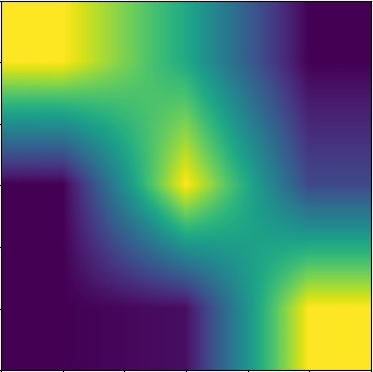}
\;\;\;\;\;\;\;\;\;
\includegraphics[width=0.6in,height=0.6in]{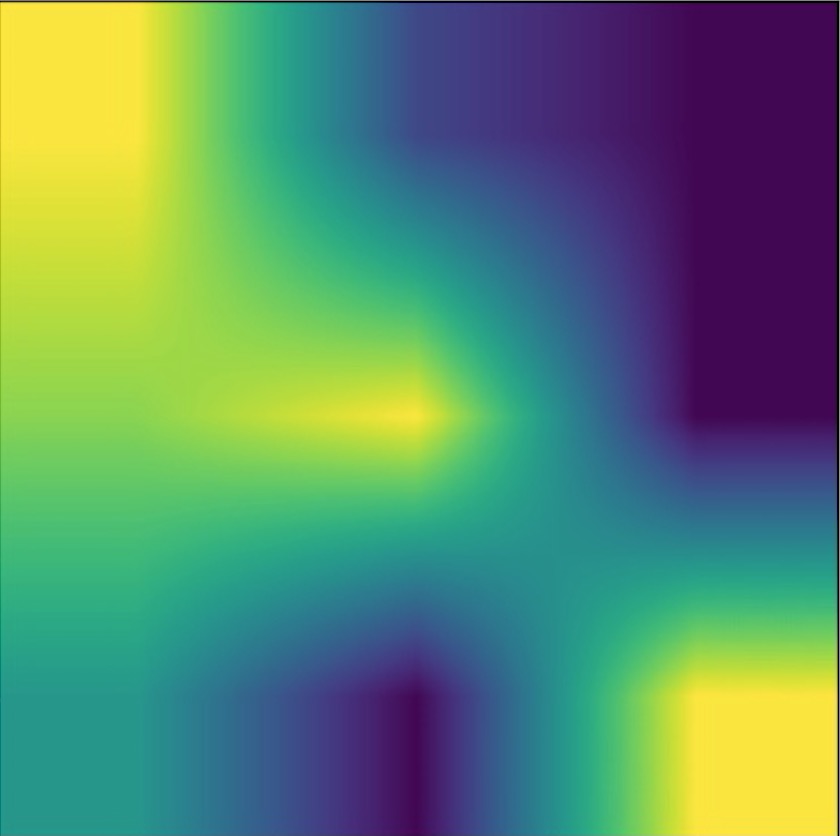}
\;\;\;\;\;\;\;\;\;
\includegraphics[width=0.6in,height=0.6in]{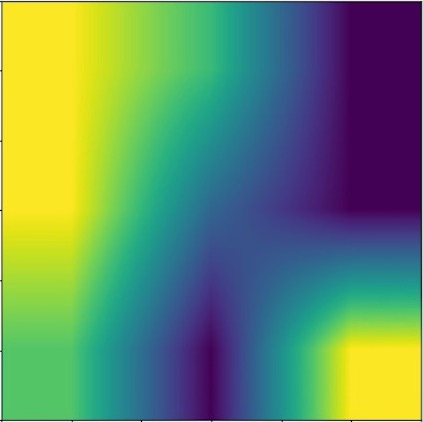}
\end{tabular}
\\  
\hline 

\begin{tabular}[c]{@{}l@{}l@{}}
video ID: 0LHWF  \\  
1. [0.0:4.1]   a person sits in a chair.\\
2. [2.1:16.2]  person holding a book.\\
3. [2.7:15.0]  person reading a book. \\
4. [2.7:15.0]  person read book. \\
\end{tabular}
&
\begin{tabular}{c} \\
\;\;\;
\includegraphics[width=0.6in,height=0.6in]{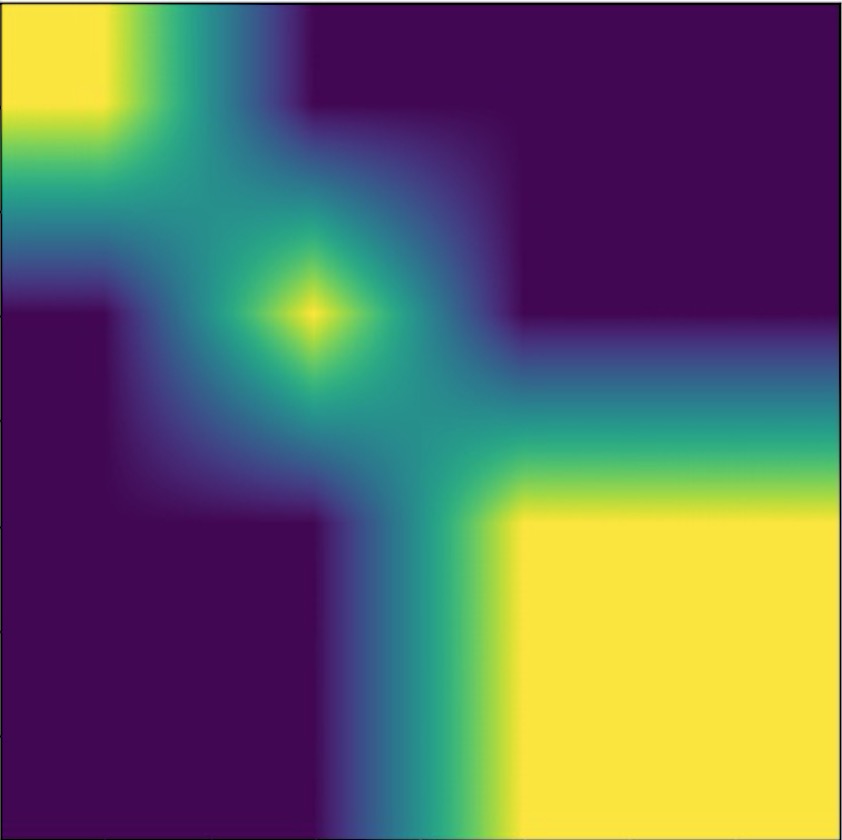} 
\;\;\;\;\;\;\;\;\;
\includegraphics[width=0.6in,height=0.6in]{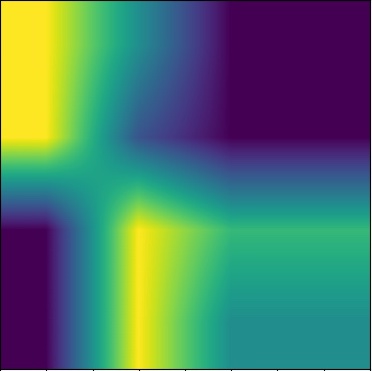}
\;\;\;\;\;\;\;\;\;
\includegraphics[width=0.6in,height=0.6in]{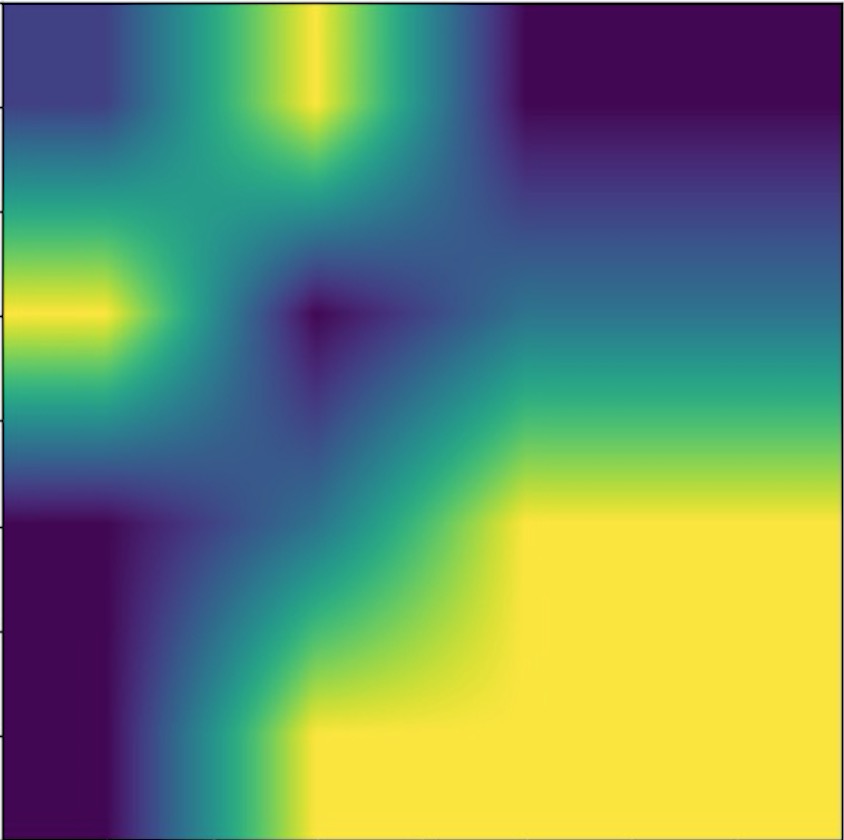}
\;\;\;\;\;\;\;\;\;
\includegraphics[width=0.6in,height=0.6in]{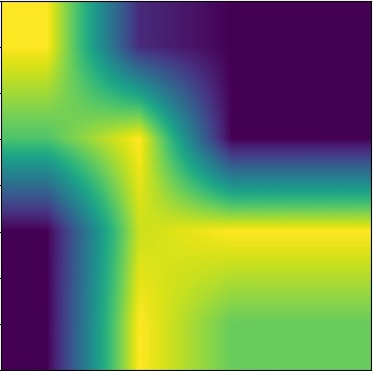}
\end{tabular}
\\  
\hline 

\end{tabular}
}
\end{center}
\caption[Visualization of similarity scores in \capt{\Earlyfusion} and \Latefusion\ models on Charades-STA dataset]
{Visualization of similarity scores between the $N$ input queries for a video and their $N$ ground truth temporal segments (resulting in a $N\times N$ confusion matrix) for the models \capt{\Earlyfusion}, \Earlyfusion\ and \Latefusion~on Charades-STA dataset.  
Warmer colors indicate higher similarity scores.
The start and end times are in seconds.
}
\label{tab:charadesSTA_videos_analysis}
\end{table*}

We evaluate our proposed models on one recent dataset designed for the text-to-clip retrieval task, Charades-STA~\cite{TALL}, and one dataset designed for the dense video caption task which has the data annotations required by the text-to-clip retrieval task, ActivityNet Captions dataset~\cite{DenseCapVid}.
We compare three versions of our model and two baselines:
\begin{itemize}
\item Random: a trivial baseline that randomly selects among among candidate clips.
\item \Latefusion{}: vector embedding approach that separately embeds the query sentence and video clip to vectors.
\item \Earlyfusion{}: our early fusion model that predicts query-clip similarity from the LSTM.
\item \queryproposal{\Earlyfusion}: LSTM with query-guided segment proposal network, as opposed to query-agnostic proposals.
\item \capt{\queryproposal{\Earlyfusion}}: our full model with the addition of captioning loss.
\end{itemize}

We follow the evaluation setup in~\cite{TALL}, 
which is adapted from a similar task in the image domain, namely, 
natural language object retrieval \cite{hu2016natural}. 
Specifically, we consider a set of temporal Intersection-Over-Union (tIoU) thresholds. 
For each threshold $\tau$, we compute the Recall@$K$ metric, defined as the fraction of sentence queries having at least one correct retrieval (having tIoU greater than $\tau$ with ground truth) in the top $K$ retrieved video clips. Following standard practice, we use $\tau\in\{0.3,0.5,0.7\}$ and $K\in\{1,5,10\}$.

\subsection{Experiments on the Charades-STA Dataset}
\label{exp:Charades-STA}
\textbf{Dataset and Setup:}
The Charades-STA dataset was introduced in \cite{TALL} for evaluating temporal localization of events in video given natural language queries. 
The original Charades dataset~\citep{sigurdsson2016hollywood} only provides a paragraph description for each video. To generate sentence-clip annotations used in the retrieval task, \citep{TALL} decomposed the original video-level descriptions into shorter sub-sentences, and performed keyword matching to assign them to temporal segments in videos.
The alignment annotations are further verified manually.
The released annotations comprise 12,408 sentence-clip pairs for training, and 3,720 for testing.

We keep all the words that appear in the training set, resulting in a vocabulary size of 1,111.
The maximum caption length is set to 10.
We sample frames at 5 fps and set the number of input frames to 768, breaking arbitrary-length input videos into 768-frame chunks, and zero-padding them if necessary.
To initialize our segment proposal network,  we finetune a  C3D model~\cite{C3D} pretrained on Sports-1M \cite{KarpathyCVPR14}, with the ground truth activity segments of 157 classes in the training videos of the Charades activity detection dataset.
We then extract proposal visual features and train the retrieval model.

\textbf{Results:}
Table~\ref{table:charades} shows the results on the text-to-clip retrieval task for Charades-STA.  
First, it is interesting to note that our vector embedding baseline (\Latefusion) already outperforms CTRL (reg-np), the best model in \cite{TALL},  by a noticeable margin. 
We believe there are two reasons for this. First, our segment proposal network from R-C3D model offers finer temporal granularity, and therefore provides cleaner  visual feature representations compared to the sliding windows approach in CTRL.
Second, we use a triplet-based loss that more effectively captures ranking constraints, compared to CTRL's binary classification loss.

Our \Earlyfusion\  model significantly outperforms the \Latefusion\ baseline.
To explore the effectiveness of the early fusion approach for fusing cross-modal features used in the retrieval model, we visualize the similarity scores (warmer colors = higher) between ground truth queries and their corresponding ground truth segments, for three example videos from Charades-STA dataset in Table~\ref{tab:charadesSTA_videos_analysis}. The ideal result is a block-diagonal matrix shown in the ``Expected" column of Table~\ref{tab:charadesSTA_videos_analysis}, which indicates that the ground truth query only has high correlation with its own ground truth segment, and can distinguish irrelevant temporal segments.
In the first example, \Latefusion\ maps two queries about the door with opposite actions ``open" and ``closing" to the same segment, while \Earlyfusion\ and \capt{\Earlyfusion} can recognize these two opposite actions on the same objects.
In the second example, \Latefusion\ confuses between the 1st and 2nd queries that are both about ``person'' and ``book", while \Earlyfusion\ starts to distinguish these two queries and \capt{\Earlyfusion} distinguishes them perfectly.
However, there are also some cases that \Latefusion\ performs slightly better as in the third example \Latefusion\ can distinguish ``a person sits in a chair." and ``person holding a book.".
In general, early fusion can capture more fine-grained action details in computing similarity scores when the objects are the same, which might be the reason for the good performance of the early fusion models.


\begin{table*}[t]
\renewcommand{\arraystretch}{1.2}
\centering
{ 
\begin{tabular}{l||ccc|ccc|ccc}
\hline
\multirow{2}{*}{Loss Weight} 
& \multicolumn{3}{|c}{tIoU=0.3} 
& \multicolumn{3}{|c}{tIoU=0.5}  
& \multicolumn{3}{|c}{tIoU=0.7}  \\
& \multirow{1}{2.4em}{\,R@1} & \multirow{1}{*}{R@5} & \multirow{1}{2.5em}{R@10} 
& \multirow{1}{2.4em}{\,R@1} & \multirow{1}{*}{R@5} & \multirow{1}{2.5em}{R@10} 
& \multirow{1}{2.4em}{\,R@1} & \multirow{1}{*}{R@5} & \multirow{1}{2.5em}{R@10} \\
\hline
$\lambda=0.5$ 
          & \textbf{52.3} & \textbf{95.3} & \textbf{99.2}
          & \textbf{34.4} & \textbf{77.0} & \textbf{92.5}  
          & \textbf{15.6} & \textbf{44.9} & \textbf{61.4} \\

$\lambda=1$  
          & {50.8} & {94.5} & {98.1} 
          & {32.5} & {76.1} & {91.2}  
          & {14.1} & {41.9} & {59.2} \\

$\lambda=2$  
          & 50.6 & 94.9 & 98.5
          & 33.5 & 76.5 & 91.3  
          & 14.3 & 43.4 & 60.3 \\
\hline
\end{tabular}}
\caption[Ablation Study of loss weight in \capt{\Earlyfusion} model on Charades-STA dataset]
{The effect of loss weight $\lambda$ in the \capt{\Earlyfusion} method, measured on the Charades-STA dataset. R@$K$ stands for Recall@$K$.
As our main task is retrieval, we consistently underweight the captioning loss with $\lambda=0.5$ in our experiments.
}
\label{table:charades_loss}
\end{table*}

Since we share parameters between two tasks in the fusion LSTM layer, \capt{\Earlyfusion} is able to further improve results on most metrics except R@5 at tIou 0.3.
Further ablations of the captioning loss weight $\lambda$ in Eq.~\ref{eq:loss_weight} for training the \capt{\Earlyfusion} method are shown in Table~\ref{table:charades_loss}. 
As our main task is retrieval, we choose $\lambda=0.5$ in our experiments.

When using the query-guided segment proposal network, the model \queryproposal{\Earlyfusion} improves on all the metrics over the early fusion model \Earlyfusion.
The layers used to incorporate the query in the query-guided SPN are cross-validated and shown in Supplement Material.
Our full model \capt{\queryproposal{\Earlyfusion}} with both query-guided proposals and captioning supervision obtains the highest results on most metrics except a minor weakness on R@5 at tIoU 0.3.

\begin{figure}[t]
\centering
\subfigure[Charades-STA retrieval examples]{
\label{fig:charades_examples}
\includegraphics[width=.94\linewidth]{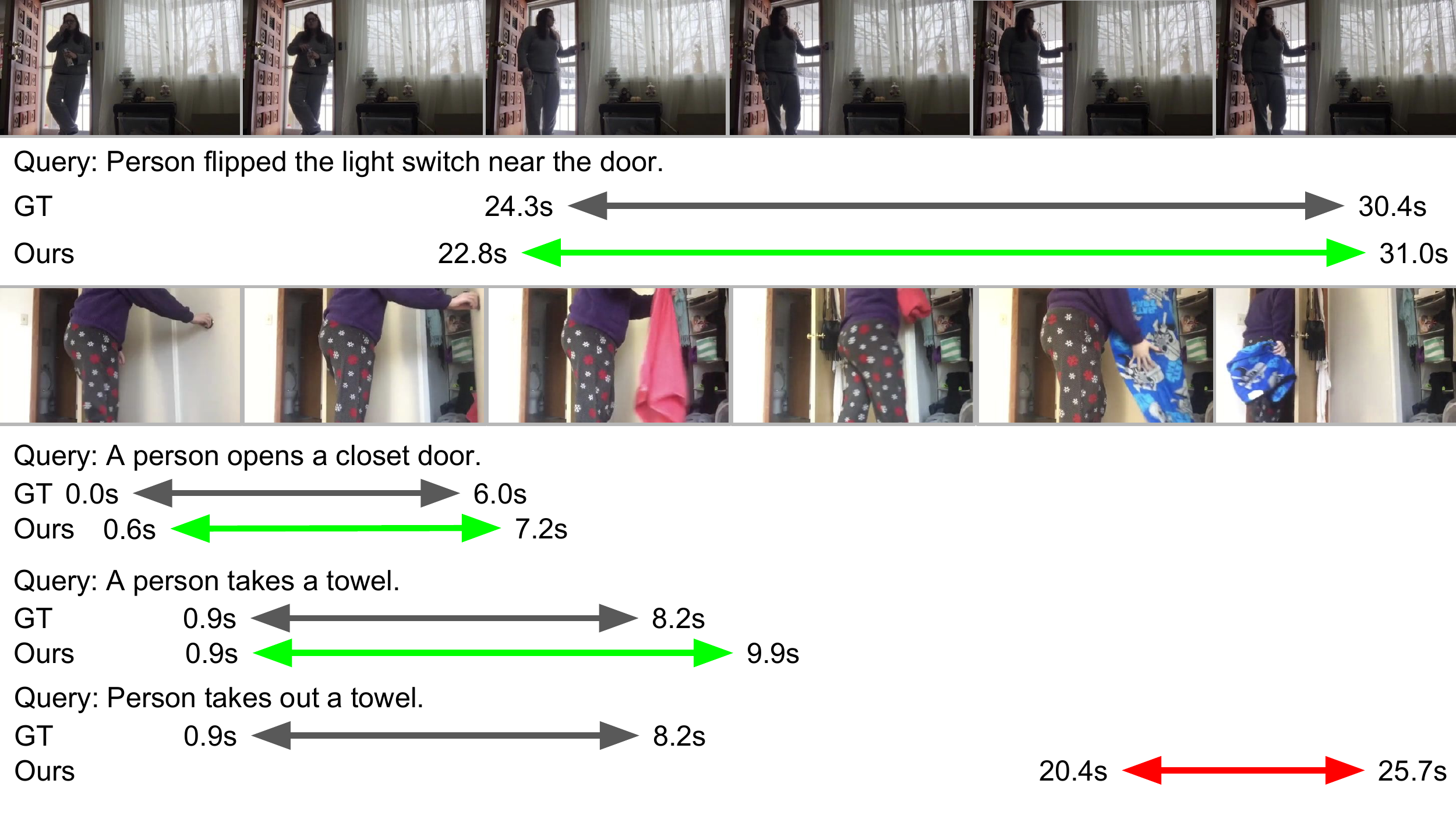}
}
\subfigure[ActivityNet Captions retrieval examples]{
\label{fig:activitynetcaption_examples}
\includegraphics[width=.94\linewidth]{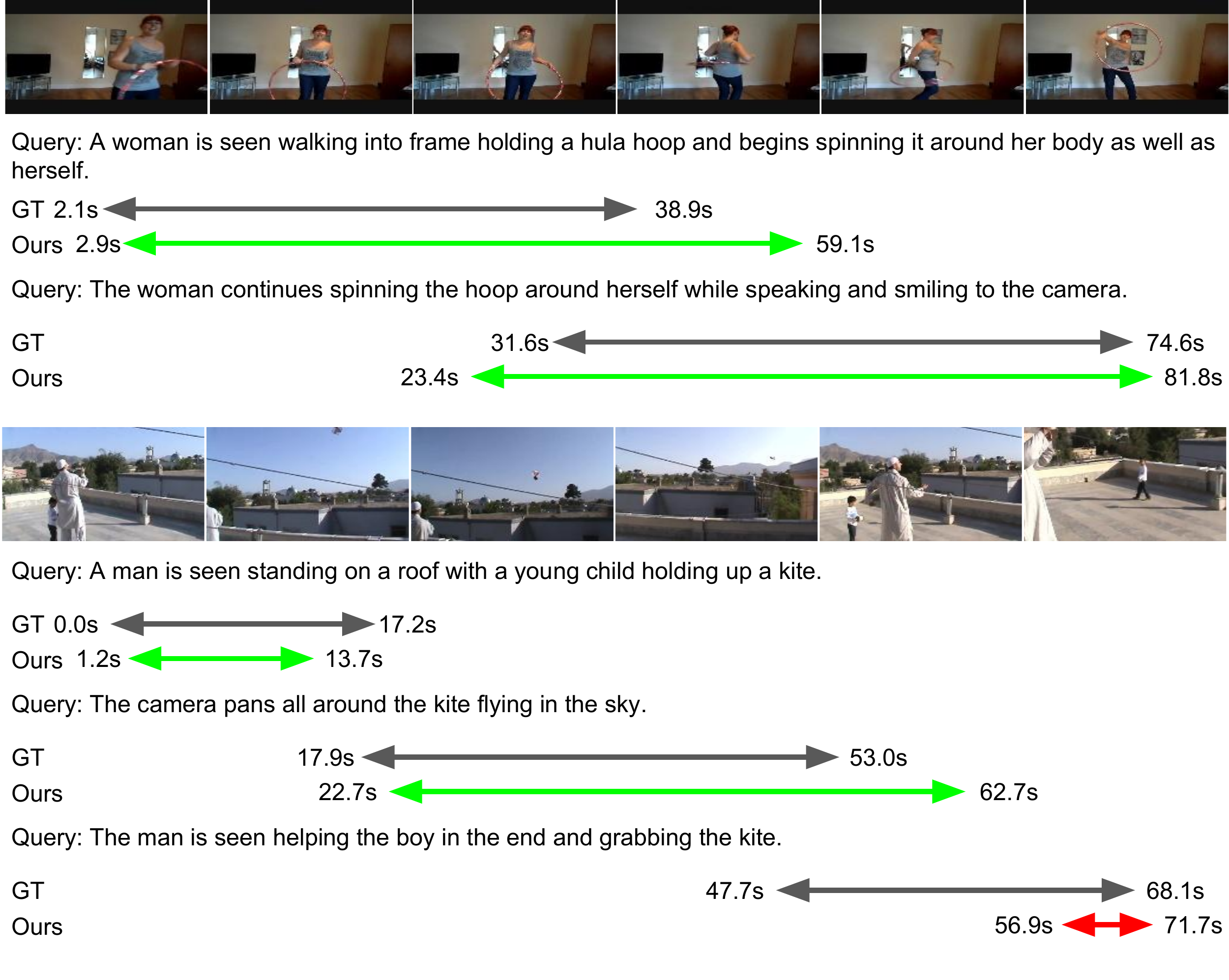}}
\caption[Qualitative visualization of text-to-clip retrieval results on the Charades-STA dataset and the ActivityNet Captions dataset]
{Qualitative visualization of example retrieval results from our full model, LSTM+QSPN+Cap, on the Charades-STA dataset~\subref{fig:charades_examples} and the ActivityNet Captions dataset~\subref{fig:activitynetcaption_examples}. 
Ground truth (GT) clips corresponding to queries are marked with black arrows. Correct predictions (predicted clips having temporal IoU more than 0.5 with ground truth)  are marked in green, and incorrect predictions are marked in red. 
The start and end times are shown in seconds. Best viewed in color.}
\label{fig:qualitative}
\end{figure}

Two example videos from the Charades-STA dataset along with query localization results are shown in Figure~\ref{fig:charades_examples}. The correct prediction is marked as green, while the wrong one is marked as red. 
Please note that the prediction is in fact correct for the query \emph{Person takes out a towel}, but is marked incorrect due to inaccurate ground truth.

\begin{table*}[t]
\renewcommand{\arraystretch}{1.2}
\centering
{
\begin{tabular}{l||ccc|ccc|ccc}
\hline
\multirow{2}{*}{Methods} 
& \multicolumn{3}{|c}{tIoU=0.3} 
& \multicolumn{3}{|c}{tIoU=0.5}  
& \multicolumn{3}{|c}{tIoU=0.7}  \\
& \multirow{1}{2.4em}{\,R@1} & \multirow{1}{*}{R@5} & \multirow{1}{2.5em}{R@10} 
& \multirow{1}{2.4em}{\,R@1} & \multirow{1}{*}{R@5} & \multirow{1}{2.5em}{R@10} 
& \multirow{1}{2.4em}{\,R@1} & \multirow{1}{*}{R@5} & \multirow{1}{2.5em}{R@10} \\
\hline
Random  
          & 5.6 & 24.8 & 42.8
          & 2.5 & 11.3 & 21.6
          & 0.8 & 4.0 & 8.1\\
\Latefusion 
          & 39.3 & 67.8 & 75.6 
          & 23.7 & 52.0 & 62.2  
          & 11.0 & 32.1 & 42.1 \\

\Earlyfusion   
          & 42.2 & 70.5 & 78.0 
          & 25.7 & 54.5 & 64.0  
          & 12.6 & 34.1 & 43.7 \\
\capt{\Earlyfusion} 
          & 42.8 & 73.5 & 80.8
          & 26.2 & 56.9 & 66.7
          & 12.6 & 35.8 & 46.3 \\
\queryproposal{\Earlyfusion}   
          & 44.2 & 74.5 & 81.9
          & 26.9 & 56.7 & 66.7
          & 13.2 & 35.5 & 45.6 \\          
\capt{\queryproposal{\Earlyfusion}}
          & \textbf{45.3} & \textbf{75.7} & \textbf{83.3} 
          & \textbf{27.7} & \textbf{59.2} & \textbf{69.3}  
          & \textbf{13.6} & \textbf{38.3} & \textbf{49.1} \\
\hline
\end{tabular}
}
\caption[Text-to-clip retrieval results on ActivityNet Captions dataset]
{Retrieval results on the ActivityNet Captions dataset~\cite{DenseCapVid}. R@$K$ stands for Recall@$K$.
Our full  model \capt{\queryproposal{\Earlyfusion}} with query-guided segment proposal network and auxiliary captioning loss outperforms all other baseline models.}
\label{table:activitynetCaption}
\end{table*}

\subsection{Experiments on ActivityNet Captions Dataset}
\label{exp:ActivityNetCaption}
\textbf{Dataset and Setup:}
The ActivityNet Captions dataset was proposed by Krishna \etal \cite{DenseCapVid} for the dense video captioning task which contains the temporal segment annotations and paired captions.
These annotations can also be used by the text-to-clip retrieval task, where the caption sentences are used as input query sentences for each video.
Each video contains at least two ground truth segments and each segment is paired with one ground truth caption.
The ActivityNet Captions dataset~\cite{DenseCapVid} contains around 20k videos and is split into training, validation and testing with a 50\%/25\%/25\% ratio.
We train on the training set and test on the combined validation sets in our experiments since the caption annotations in the test set are withheld for challenge purpose.

We keep the words in the training set with frequency greater than five to build a vocabulary of size 3,892, and set the maximum caption length to 30.
We sample frames at 3 fps, and set the maximum number of input frames in the buffer to be 768.
Again, a  3D ConvNet model~\cite{C3D} pretrained on the Sports-1M dataset and finetuned on ground  truth  activity  segments of ActivityNet detection dataset is used to initialize our segment proposal network. 
We use the good settings from the ablation studies on the Charades-STA dataset.

\textbf{Results:}
Results using the standard evaluation protocol are given in Table~\ref{table:activitynetCaption}. 
Similar trends can be observed for the five variants of our model, as in the Charades-STA experiments.
\Earlyfusion\ significantly outperforms the baseline \Latefusion.
Also, with the assistance of the captioning loss,  the multi-task model \capt{\Earlyfusion} does better than \Earlyfusion.
With input query sentences regulating the segment proposal network, all the metrics in the model~\queryproposal{\Earlyfusion} improve compared to using the original segment proposal network in \Earlyfusion.
Our full model~\capt{\queryproposal{\Earlyfusion}} with both captioning supervision and query-guided proposals gets highest results in all the metrics.


Two example retrieval results from the ActivityNet Captions dataset can be found  in Figure~\ref{fig:activitynetcaption_examples}. In the first example, our model localizes the precise moment described by the query sentences about playing hula hoop. 
In the second example, it also correctly identifies the events corresponding to the queries \textit{A man is seen standing on a roof with a young child holding up a kite} and \textit{The camera pans all around the kite flying in the sky}.
However, for the third query \textit{The man is seen helping the boy in the end and grabbing the kite}, our model directly localizes to an incorrect shot segment at the very end of the video, maybe biased by the phrase ``in the end" in the input query.

\section{Conclusion}
\label{sec:conclusion}

In this paper, we address the problem of text-to-clip retrieval: temporal localization of events within videos that match a given natural language query. 
We introduce a multilevel feature integration model to fuse language and vision earlier and more tightly than existing methods, which are typically based on the late fusion approach of learning independent vector embeddings.
We make use of a segment proposal network to filter out unlikely clips, and improve it by conditioning on the input query.
We learn a two-layer LSTM to directly predict similarity scores between sentence queries and video clips, and augment its training loss by adding the captioning task.

Evaluated on two challenging datasets, our approach performs more accurately than previous methods when retrieving clips from many possible candidates in untrimmed videos. 
For future work, 
we are interested in further exploiting language features in order to modulate the extraction of visual features, similar to \cite{NIPS2017_vries}.
\smallskip

\noindent{\bf Acknowledgements:} Supported in part by IARPA (contract number D17PC00344) and DARPA's XAI program.

{\footnotesize The U.S. Government is authorized to reproduce and distribute reprints for Governmental purposes notwithstanding any copyright annotation thereon. Disclaimer: The views and conclusions contained herein are those of the authors and should not be interpreted as necessarily representing the official policies or endorsements, either expressed or implied, of IARPA, DOI/IBC, or the U.S. Government.}

\bibliography{6_bib}
\bibliographystyle{aaai}
\end{document}